%% file: acl2023.tex
\pdfoutput=1

\documentclass[11pt]{article}

\usepackage[]{ACL2023}

\usepackage{times}
\usepackage{latexsym}

\usepackage[T1]{fontenc}

\usepackage[utf8]{inputenc}

\usepackage{microtype}

\usepackage{inconsolata}
\usepackage{bbm}
\usepackage{times}
\usepackage{amsmath}
\usepackage{amsthm}
\usepackage{booktabs}
\usepackage{multirow}
\usepackage{latexsym}
\usepackage{subcaption}
\usepackage{graphicx}
\usepackage{hyperref}
\usepackage[T1]{fontenc}
\usepackage{url}
\usepackage[utf8]{inputenc}
\usepackage{tikz}
\usetikzlibrary{shapes.geometric, arrows}
\definecolor{mygray}{rgb}{0.86,0.86,0.86}
\usepackage{inconsolata}
\usepackage{enumitem}
\usepackage{appendix}

\definecolor{myorange}{HTML}{ff7f0e}
\definecolor{myblue}{HTML}{1f77b4}
\definecolor{mygreen}{HTML}{2ca02c}
\tikzstyle{io} = [trapezium, trapezium left angle=70, trapezium right angle=110, minimum width=1.5cm, minimum height=0.5cm, text centered, draw=myblue, line width=1.5pt]
\tikzstyle{process} = [rectangle, minimum width=1.5cm, minimum height=0.5cm, text centered, draw=myorange, line width=1.5pt]
\tikzstyle{decision} = [diamond, minimum width=1.5cm, minimum height=0.5cm, text centered, draw=mygreen, line width=1.5pt]
\tikzstyle{arrow} = [thick,->,>=stealth]

\theoremstyle{plain}

\usepackage[export]{adjustbox}
\usepackage[utf8]{inputenc}

\usepackage{float}
\newfloat{algorithm}{t}{lop}
\usepackage{adjustbox}

\title{Semantic Sensitivities and Inconsistent Predictions: Measuring the Fragility of NLI Models}

\usepackage{lipsum}

\newcommand\blfootnote[1]{%
  \begingroup
  \renewcommand\thefootnote{}\footnote{#1}%
  \addtocounter{footnote}{-1}%
  \endgroup
}
\makeatother

\author{Erik Arakelyan$^{\dagger}$, Zhaoqi Liu$^{\dagger}$, Isabelle Augenstein$^{*}$ \\
Department of Computer Science\\
University of Copenhagen \\
Copenhagen Denmark  \\ 
\texttt{\{erik.a,nbq899,augenstein\}@di.ku.dk}
}

\begin{document}
\maketitle

\blfootnote{$^{\dagger}$Equal contribution, alphabetical order. \ $^*$Senior author.}

\begin{abstract}

Recent studies of the emergent capabilities of transformer-based Natural Language Understanding (NLU) models have indicated that they have an understanding of lexical and compositional semantics.
We provide evidence that suggests these claims should be taken with a grain of salt: we find that state-of-the-art Natural Language Inference (NLI) models are sensitive towards minor semantics preserving surface-form variations, which lead to sizable inconsistent model decisions during inference. 
Notably, this behaviour differs from valid and in-depth comprehension of compositional semantics, however does neither emerge when evaluating model accuracy on standard benchmarks nor when probing for syntactic, monotonic, and logically robust reasoning.
We propose a novel framework to measure the extent of semantic sensitivity. 
To this end, we evaluate NLI models on adversarially generated examples containing minor semantics-preserving surface-form input noise. This is achieved using conditional text generation, with the explicit condition that the NLI model predicts the relationship between the original and adversarial inputs as a symmetric equivalence entailment.
%
%
We systematically study the effects of the phenomenon across NLI models for \emph{in-} and \emph{out-of} domain settings. 
Our experiments show that semantic sensitivity causes performance degradations of $12.92\%$ and $23.71\%$ average over \emph{in-} and \emph{out-of-} domain settings, respectively. 
We further perform ablation studies, analysing this phenomenon across models, datasets, and variations in inference and show that semantic sensitivity can lead to major inconsistency within model predictions.

\end{abstract}

\section{Introduction}

Transformer-based \citep{vaswani2017attention} Language Models (LMs) have shown solid performance across various NLU tasks \citep{wang2018glue, wang2019superglue}. These advances have led to suggestions regarding the emergent capabilities of the models in terms of syntactic \citep{sinha2020unnatural,hewitt2019structural, jawahar2019does, warstadt2020can}, logic \citep{wei2022emergent,wei2022chain} and semantic \citep{kojima2022large,dasgupta2022language} understanding. However, we present novel evidence that indicates that these models are prone to inconsistent predictions induced by inherent susceptibility towards semantic sensitivities.

To probe the models for these discrepancies, we formalise \emph{semantic comprehension} as the ability to distinguish logical relations within sentences through identifying compositional semantics \citep{jacobson2014compositional,carnap1959introduction}. This means that negligible semantic variations should not impact the inherent relations implied between the texts, e.g. \emph{``There were beads of perspiration on his brow.''} entails both \emph{``Sweat built up upon his face.''} and the slight variation \emph{``The sweat had built up on his face.''}
Authentic comprehension of semantics does allow for such understanding through discovering semantic structures and the inherent relations induced by them \citep{cicourel1991semantics,schiffer1986compositional,rommers2013context}. This means that analysing the emergent semantic understanding within a model should minimally involve testing for sensitivity towards semantics-preserving surface-form variations.

\begin{figure*}[t]
    \centering
    \includegraphics[width=\textwidth]{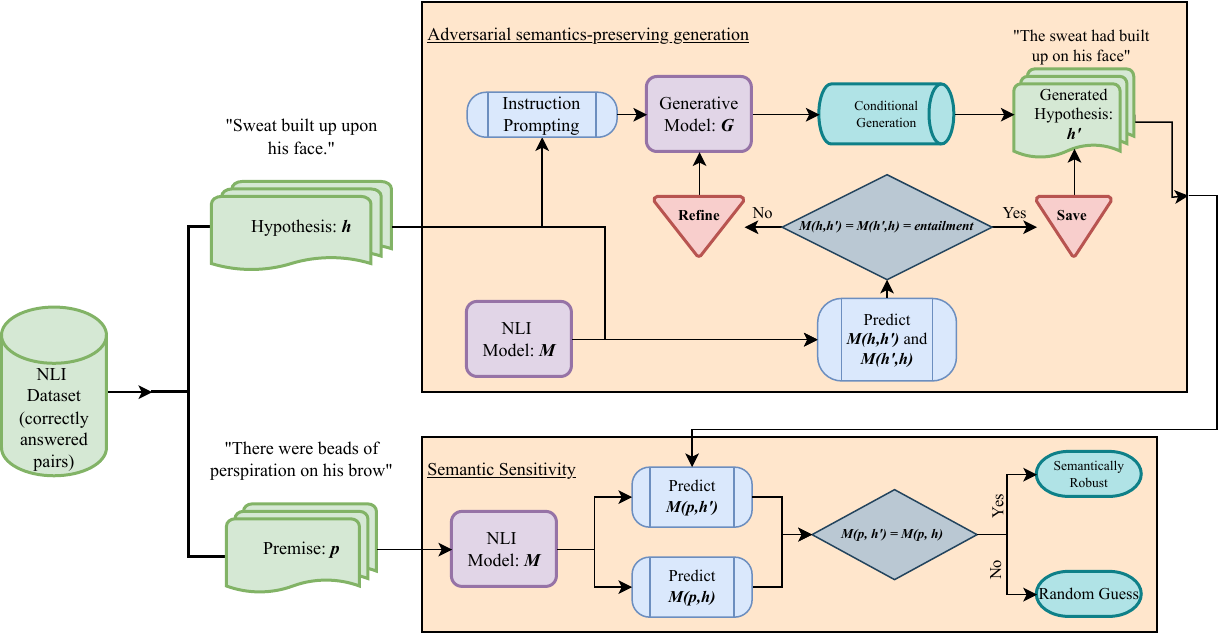}
    \caption{The proposed framework is comprised of two components. (i) a module for generating semantics-preserving surface-form hypothesis variations and (ii) using the generated surface for measuring semantic sensitivity and predictive inconsistency.}
    \label{fig:framework}
\end{figure*}

We particularly focus on the task of textual entailment \citep{dagan2005pascal}, otherwise referred to as Natural Language Inference \citep[NLI]{ bowman2015large}, which has been widely used to probe how well the models understand language \citep{condoravdi2003entailment, williams2017broad, nie2019adversarial}. This is a pairwise input task, where given a premise $p$ and a hypothesis $h$, the objective is to predict if the premise \emph{entails, contradicts} or is \emph{neutral} towards the hypothesis.

We propose a framework for testing semantic sensitivity within transformer-based models trained for NLI, by creating semantics-preserving surface-form variations of the hypothesis (see \autoref{fig:framework}).
These variations are created using conditional generation with Large Language Models (LLMs). We show that proposed candidates do not alter the core meaning or the truth value compared to the original statement. The original and generated sentences maintain denotative equivalence, where two sentences or phrases might be interpreted as having the same truth value or factual content but may carry minor variations of nuances or connotations.
To ensure that the relations are preserved within the candidates during conditional generation, we assert that the NLI model predicts the original and generated hypothesis to symmetrically entail each other. This indicates that the model perceives both the generated and original hypothesis as equivalent. 
After introducing these variations, we evaluate the NLI model by replacing the original hypothesis with the generated candidates. As the candidates are indicated to be equivalent by the same NLI model, this evaluation will indicate whether the model can recover the existent relation between the premise hypothesis pair in the presence of minor semantic-preserving noise. 
We use the samples where the model identifies the existing relation correctly from the original premise hypothesis pair. This ensures that assessing for semantic sensitivity would not be hindered by the discrepancies in model performance.

We systematically study the semantic sensitivity across transformers that achieve state-of-the-art or similar results when trained on NLI datasets, namely RoBERTa \citep{liu2019roberta}, BART \citep{lewis2019bart}, DeBERTa \citep{he2020deberta} and DistilBart \citep{sanh2019distilbert, lewis2019bart} with different parametrizations. To measure the effect of the phenomenon on the inconsistency of the predictions, we use three popular English datasets - MultiNLI \citep[MNLI]{williams2017broad}, SNLI \citep{bowman2015large} and ANLI \citep{nie2019adversarial}. The models are fine-tuned using MNLI, which we choose for \emph{in-domain} testing, as it covers a wide range of topics and is frequently used for zero-shot and few-shot textual classification \citep{yin2019benchmarking}. We use the same models for \emph{out-of-domain} evaluation across the other NLI datasets.

Our contributions are as follows: (i) we propose a novel framework for assessing semantic sensitivity within transformer-based language models (ii) we systematically study the influence of this phenomenon on inconsistent predictions across various transformer variants (iii) we show that the effect is persistent and pronounced across both \emph{in-} and \emph{out-of-domain} evaluations (iv) we further complete ablations to assess the severity of the inconsistent predictions caused by semantic sensitivity.

\section{Related Work}

Semantic comprehension is considered a fundamental building block for language understanding \citep{allen1995natural}. Although attempts have been made to probe language models in terms of compositional semantic capabilities, the conclusions regarding their emergence remain to be discussed.

\paragraph{Models appear to understand semantics}

Recently a wide suite of tasks has been proposed for testing models for language understanding \citep{wang2019superglue,zellers2018swag, ribeiro2020beyond} with the credence that a model with strong performance should be able to utilise semantic relations when completing the tasks. In light of these, it has been shown that transformer-based language models can be directly trained \citep{zhang2020semantics, rosset2020knowledge} to utilise semantic structure to gain distributional information within the task. Specifically, NLI models have also been shown to be capable of pragmatic inferences \citep{jeretic2020natural} with a perception of implicature \citep{grice1975logic} and presupposition \citep{stalnaker1977pragmatic,grice1975logic}.

\paragraph{Models struggle with semantics}

Directly probing for a specific aspect of semantic understanding has shown that transformer-based language models tend to struggle with semantics \citep{belinkov2022probing}. It has been indicated that pretraining the language models does not exploit semantic information for entity labeling and coreference resolution \citep{liu2019linguistic}. Furthermore, transformer attention heads only minimally capture semantic relations \citep{kovaleva2019revealing} from FrameNet \citep{baker1998berkeley}. Studies have also shown that NLI models, in particular, tend to struggle with lexical variations, including word replacements \citep{glockner-etal-2018-breaking, ivan2018behavior, geiger2020neural}, and sequence permutations \citep{sinha2021unnatural}.

\input{tables/nli_eval}

\paragraph{Sensitivity in NLI models}

Probing NLI models for language understanding has been a hallmark testing ground for measuring their emerging capabilities \citep{naik2018stress, wang2015learning, williams2017broad}. A wide range of tests indicates that models trained for NLI are prone to struggling with syntax and linguistic phenomena \citep{dasgupta2018evaluating, naik-etal-2018-stress, an-etal-2019-representation, ravichander-etal-2019-equate, jeretic-etal-2020-natural}. It has also been shown that NLI models heavily rely on lexical overlaps \citep{ivan2018behavior, mccoy-etal-2019-right, naik-etal-2018-stress} and are susceptible to over-attending to particular words for prediction \citep{gururangan-etal-2018-annotation, clark-etal-2019-bert}.
Our line of work is associated with evaluating NLI models for monotonicity reasoning \citep{yanaka2019can} and sensitivity towards specific semantic phenomenon \citep{richardson2020probing}, such as boolean coordination, quantification, etc. However, we systematically test NLI models for their compositional semantic abilities and measuring the degree of inconsistence of their predictions influenced by the phenomenon.

\section{Methodology}

We aim to create a framework for assessing semantic sensitivity within NLI models and measure its impact on the inconsistence of model predictions. The first part of the pipeline we propose is an adversarial semantics-preserving generation for introducing variations within the original samples. The second part of the pipeline involves assessment using the acquired generations.

\subsection{Semantics Preserving Surface-Form Variations}

We formalise NLI as a pairwise input classification task. Given a dataset of premise hypothesis pairs $\mathcal{D} = {(p_1,h_1), \dots (p_n,h_n)}$, where ${ \forall p_i \in P \And h_i \in H}$ are a set of textual tokens $P,H \subseteq \mathcal{T}$, the goal is to classify the pairs as \emph{entailment, contradiction} or \emph{neutrality}, i.e. $\mathcal{C} = \{ E,C,N\}$. We are also given a pre-trained language model (PLM) $\mathcal{M}$ that is trained for textual entailment.
Before introducing semantic variations, only the samples where model $\mathcal{M}$ predicted the label correctly are filtered, i.e. $D_\text{correct} = \{ \forall (p_i,h_i)\in \mathcal{D} : \mathcal{M}(p_i,h_i) = \hat{y} = y\}$, where $\hat{y}$ is the prediction and $y$ is the original label. This is completed to ensure that the evaluation of semantic sensitivity is not hindered or inflated by the predictive performance and confidence of the model $\mathcal{M}$. This type of filtering is used when probing for emergent syntactic \citep{sinha2021unnatural}, lexical \citep{jeretic-etal-2020-natural}, and numerical \citep{wallace2019nlp} reasoning capabilities. We can see the original accuracy of NLI models and the number of samples used in the study in \autoref{tab:nli_eval}.

To introduce semantics preserving noise within chosen samples, we complete a two-fold refinement process. We utilise a generative LLM $\mathcal{G}$, which has been fine-tuned on natural language instructions \citep{wei2021finetuned, chung2022scaling}, and prompt it to paraphrase the original hypothesis $h_i$, with the following prompt: \emph{Rephrase the following sentence while preserving its original meaning: <$h_i$>. }
This is not sufficient to produce semantics-preserving variations as generative models are prone to hallucinations \citep{ji2023survey} and not assured to produce an equivalent paraphrase. To ensure that the generation $h_i^{\prime}$ is logically equivalent to the original sample and thus semantics-preserving, we impose the condition that the NLI model should infer the relation between the original and generated hypothesis as a symmetric entailment:
\begin{align}\label{eq:label_change}
    \mathcal{M}(h_i, h_i^{\prime}) = \hat{y}_{\mathcal{C}=E} = \mathcal{M}(h_i^{\prime}, h) 
\end{align}

The bidirectional nature of entailment allows us to claim that sentences are logically equivalent \citep{angell1989deducibility,clark1967general}. We refine the proposed variation candidates using the generator $\mathcal{G}$ until $k$ candidates that satisfy the condition are produced.

\paragraph{Human Evaluation of Surface-Form Variations}

To further ensure the validity of this variation generation method, we conduct a human evaluation of the generated samples. We randomly sample $100$ examples of generated and original hypothesis pairs across all datasets and employ two annotators to assess whether the sentences are semantically and logically equivalent within the pair. Our results show that in $99\%$ of the cases, the annotators marked the samples as equivalent with an inter-annotator agreement measure of Cohen's $\kappa = 0.94$. This further shows the reliability of the method for generating semantics-preserving surface form variations. We provide further token overlap level analysis in \autoref{sec:appendix}.


\input{tables/fooling_rate}

\subsection{Evaluating Semantic Sensitivity}

After obtaining $k$ semantic variations for each hypothesis, we test the semantic sensitivity of the model by replacing the original hypothesis $h_i$ with the candidates $\{h_i^{\prime 1}, \dots h_i^{\prime k} \}$ and making a prediction with the NLI model $\mathcal{M}$. As the proposed variations are logically equivalent to the original, we want to test if the new model prediction would vary compared to the original.

\begin{multline}
\mathcal{R}(p_i, h_i, h_i^{\prime j}, \mathcal{O}) = \\
    = \begin{cases}
       1, \mathcal{O}(\mathcal{M}(p_i, h_i), \mathcal{M}(p_i, h_i^{\prime j})) = 0 \\
       0, \mathcal{O}(\mathcal{M}(p_i, h_i), \mathcal{M}(p_i, h_i^{\prime j})) = 1
   \end{cases}
\end{multline}

Here $\mathcal{O}: \mathcal{C} \times \mathcal{C} \rightarrow \{0, 1\}$ is a boolean matching operator between the labels predicted with original hypothesis $h_i$ and the surface-form variations $h_i^{\prime j}$.
A change in the label would imply that the model is semantically sensitive and the original correct prediction is inconsistent with the label produced for the semantics preserving surface-form variation.  
A graphical representation can be seen in \autoref{fig:semantic_triangle}.
We use two metrics to measure semantic sensitivity within NLI models, both of which are derivative formulations of a Fooling Rate \citep{moosavi2017universal}, which is used for assessing the success of adversarial attacks \citep{chakraborty2018adversarial}. Given $k$ possible surface-form variations for the hypothesis, we test if at least one of the candidates would be able to cause a label change compared to the original prediction, which can be formalised as:

\begin{align}\label{eq:soft_f_r}
    r_r=\frac{\sum_i^{n^\prime}\mathbbm{1}\left[\exists j \in [1,k], \mathcal{R}(p_i, h_i, h_i^{\prime j}, =) \neq 1\right]}{n^\prime}.
\end{align}

Here $n^\prime$ is the number of correctly answered original samples, and the matching operator $\mathcal{O}$ is a simple equality checking operator "$=$". We refer to this metric as a relaxed Fooling Rate. To measure more drastic label changes, i.e. \emph{entailment} to \emph{contradiction} and vice versa, we also define a stricter version of \autoref{eq:soft_f_r}.

\begin{align}\label{eq:strict_f_r}
    r_s=\frac{\sum_i^{n^\prime}\mathbbm{1}\left[\exists j \in [1,k], \mathcal{R}(p_i, h_i, h_i^{\prime j}, =^s) \neq 1\right]}{n^\prime}.
\end{align}

 We replace standard equality for the operator $\mathcal{O}$ in \autoref{eq:soft_f_r} with a strict counterpart that matches only if the predictions are direct opposites, i.e. \emph{entailment} $\leftrightarrow$ \emph{contradiction}. It must be noted that the \emph{neutral} class does not have a direct opposite; thus, the metric for this label remains unchanged. It can be concluded that the inequality $r_s \leq r_r \leq 1$ trivially holds when using these metrics.

\section{Experimental Setup}

\subsection{Model Details}

\paragraph{Semantics preserving Generation}

To generate and refine semantic variations of the original hypothesis, we chose \emph{flan-t5-xl} as the generation model $\mathcal{G}$. It is an instruction-tuned LLM that has shown close state-of-the-art performance in tasks such as paraphrasing, zero and few shot generation, chain of thought reasoning (CoT), and multi-task language understanding \citep{chung2022scaling}. For each of the selected hypotheses, we produce $k=5$ unique semantics-preserving variations. To ensure diversity and consistency of the generated text while avoiding computationally expensive exhaustive search, we use a group beam search \citep{vijayakumar2016diverse}  with a temperature $t\in [0.3, 0.6]$ and a maximum output of 40 tokens throughout the generation and refinement procedure. We also further diversify the generation by using the recipe from \citet{li2016simple}.

\paragraph{NLI models}

We systematically experiment with transformer architectures that are fine-tuned on MNLI, which exhibit state-of-the-art or close predictive accuracy on the dataset. We specifically choose \emph{bart-large} \citep{lewis2019bart}, \emph{roberta-large} \citep{liu2019roberta}, \emph{deberta-base, deberta-large, deberta-xlarge} \citep{he2020deberta} and \emph{distilbart} \citep{sanh2019distilbert}. These PLMs are taken without change from their original studies through the Transformers library \citep{wolf-etal-2020-transformers}, ensuring the complete reproducibility of the results. To observe the effect in an \emph{out-of-domain} setup, we also evaluate these models on SNLI and ANLI in a zero-shot transfer setting.

\section{Results and Analysis}

This section presents the results and analyses of our semantic sensitivity evaluation framework along with a suite of ablations analysing the phenomenon across various transformer sizes, domains, and label space. Furthermore, we measure the impact of the phenomenon on the inconsistent predictive behaviour of NLI models. 

\subsection{Semantic Sensitivity}

\paragraph{In-domain}

We evaluate several PLMs trained on MNLI using our experiments presented in \autoref{tab:foolrate1}. The results show that models are limited in their comprehension of compositional semantics as the relaxed fooling rate on \emph{in-domain} experimentation averages at $r_r = 12.9\%$. This is further reinforced by the fact that more than half, $r_s = 6.58\%$  of the label changes occur with strict inequality. This means that minor semantics-preserving changes lead to a sizable shift in model predictions, even prompting towards the opposite decision edge half the time. The behaviour is consistent across all the transformers and leads us to believe that samples that changed labels after surface-form variations showcase the inconsistent predictive nature of the models. We further elaborate on this in the next section.
Consequently, semantically equivalent variations evidently hinder the decision-making of the NLI models, prompting us to believe that models have limited understanding w.r.t. semantic structure and logical relation, even when the model is trained on texts from the same distribution.

\begin{figure*}[t]
    \centering
    \includegraphics[width=0.8\textwidth]{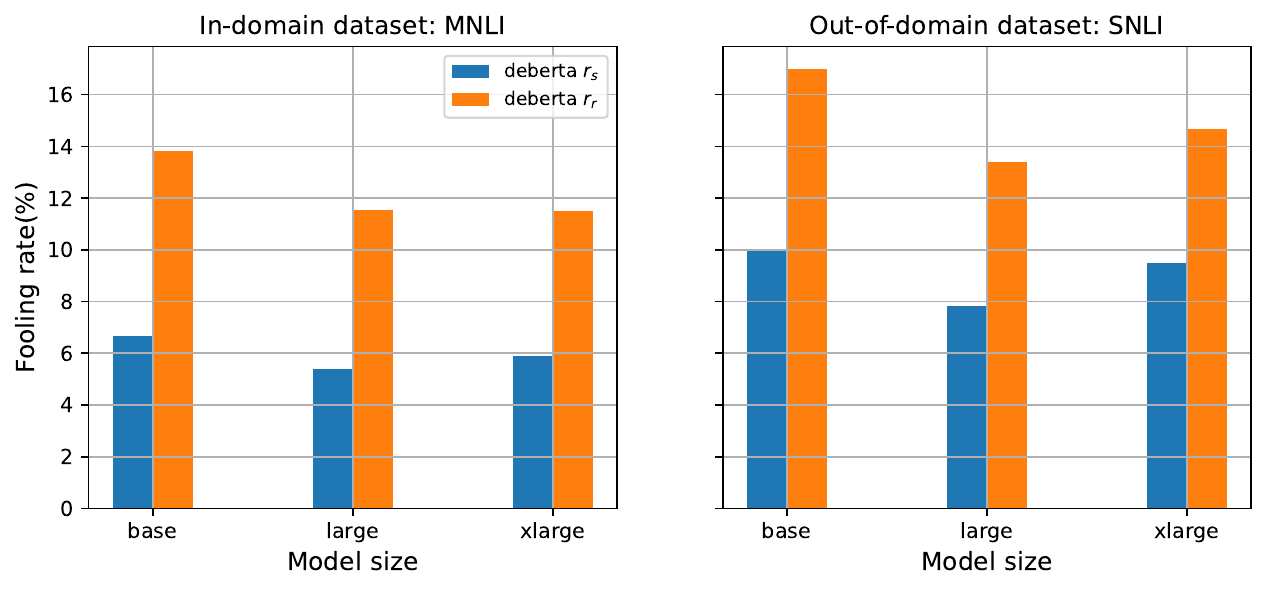}
    \caption{In- and out-of-domain fooling rate of DeBERTa of varied sizes, which are measured on MNLI (left) and SNLI (right). Similarly, $r_s$ and $r_r$ represent the strict and relaxed fooling rates, respectively.}
    \label{fig:frvary}
\end{figure*}

\paragraph{Out-of-domain}

We also probe the NLI models in an \emph{out-of-domain} zero-shot setting to assess the transferability of compositional semantic knowledge. Our results in \autoref{tab:foolrate1} show that the discrepancies and limitations in semantic comprehension are even more pronounced in this setting. We see an averaged relaxed fooling rate of $r_r = 23.7\%$, with the maximum at $57.49\%$, which is only marginally better than a majority voting baseline.
It must be noted that because different datasets have varying numbers of samples, the average is weighted w.r.t. the number of sampled instances from the particular dataset in the experiment.
The results on \emph{out-of-domain} evaluation once again follow the pattern that more than half, $r_s = 15.8\%$ of the samples switch the labels to their logically contrasting counterparts. This shows that zero-shot transfer further amplifies the limitations that NLI models have for using semantic structures and preserving logical relations. This further suggests that the semantic variations where a label change occurs are likely to be originally predicted correctly as an inconsistent guess. It follows, that although PLMs fine-tuned on MNLI are widely used for zero-shot classification, their effectiveness diminishes if the classification tasks require syntactic understanding. Indeed, model effectiveness declines and the fooling rates rise as the tasks become more challenging, requiring greater syntactic knowledge, as we can see from the comparison of the results from SNLI to ANLI.

\paragraph{Effects of distillation}

Next, we want to probe if the susceptibility towards semantic noise is transferred during model distillation. Thus, we use \emph{DistilBart} that is distilled from a larger pre-trained BART model. While model accuracy remains comparable to the original model in \autoref{tab:nli_eval}, the distilled version struggles sizeably more with surface-form variations. On average, across \emph{in-} and \emph{out-of-} domain evaluation, the distilled NLI model is more sensitive than the original in terms of relaxed fooling rate by $\triangle r_r = 18.4\%$. The effect of supposed inconsistence is amplified when observing the strict fooling rate, where on average $\frac{r_r}{r_s} \leq 1.5$. This indicates that during distillation, models are bound to forget the knowledge regarding compositional semantics making it harder to preserve the logical equivalence during inference.

\paragraph{Effects of model size}

We also test how semantics-preserving noise affects models of different sizes and parametrization (see \autoref{fig:frvary}). Although for \emph{in-domain} setup, the relaxed fooling rate metrics marginally drop as the models get bigger, the same cannot be observed in \emph{out-of-domain} setup. It is evident that bigger PLMs from our study are almost as restricted in semantic comprehension as their smaller counterparts. This indicates that emergent semantic capabilities are not only tied to model size, but also widely depend upon the choice of the training dataset.

\input{tables/fooling_rate_class}

\subsection{Severity of Inconsistent Predictions}

\paragraph{Consistency across label space}

To analyse the extent of semantic sensitivities within NLI models we test the effect across all the classes in the label spaces, presented in \autoref{tab:foolrate_class}. The per-class breakdown of the strict and relaxed fooling rate indicates that the effect is consistent across the whole label space. This allows us to conclude that the observed limitations in compositional semantic understanding are not caused by class imbalances and are not specific to a particular set of examples. We see the increased fooling rate across all of the labels when comparing \emph{in-domain} and \emph{out-of-domain} experiments. This reinforces the prior indications regarding models' inability to use semantic structure to preserve inherent relations within the data, as all logical relations attain rather similar amounts of fooling rate during direct evaluation.

\begin{figure*}[t]
    \centering
    \includegraphics[width=0.85\textwidth]{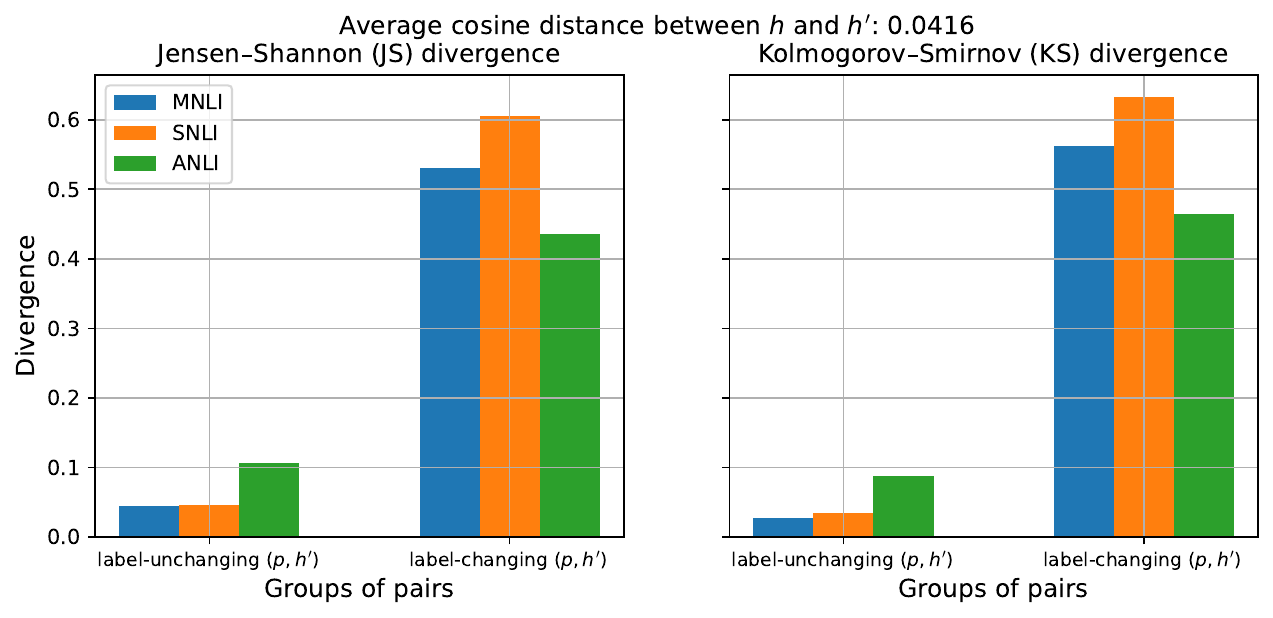}
    \caption{Divergence of predictive probability distribution between $(p,h)$ and $(p,h^\prime)$ measured across the datasets (ANLI is averaged over the rounds) and averaged over all models. All evaluation pairs are split into two groups based on whether they manage to flip the original label. Two divergence metrics are shown -- JS divergence (left) and KS divergence (right).}
    \label{fig:div}
\end{figure*}

\begin{figure}[t]
    \centering
    \includegraphics[width=\columnwidth]{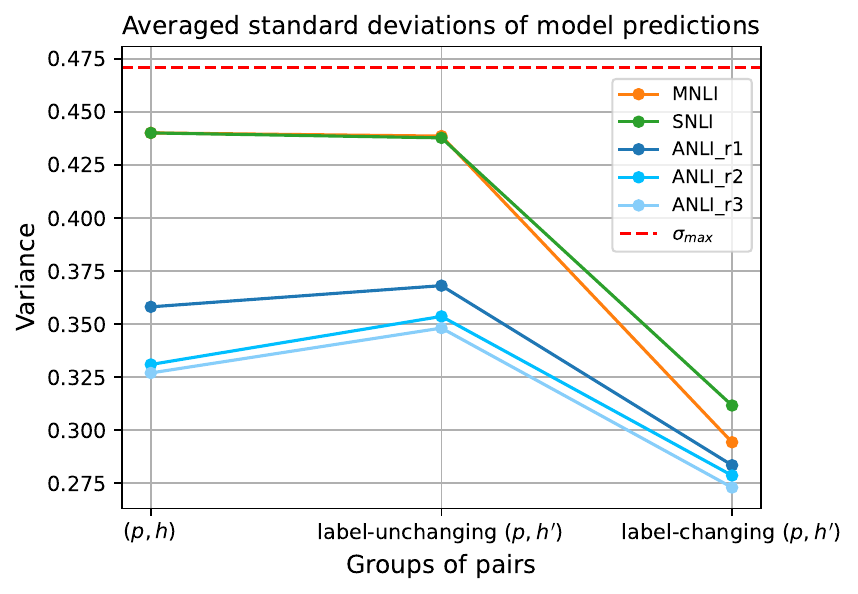}
    \caption{Standard deviation $\sigma$ of predicted label probabilities (obtained from the final softmax layer of the model) averaged for original premise-hypothesis pair (left), surface-form variations that did not cause label changes (mid) and did induce label change (right). The bigger $\sigma$, the more confident the model is w.r.t. the predictions. The results are averaged over all models.}
    \label{fig:var}
\end{figure}

\paragraph{Distribution shift in decision making}

Recall that we want to measure the impact of semantics-preserving surface-form variations on NLI models. We study the predictive distributional shift within the samples that cause a changed model prediction. To do this, we initially split the samples into two categories considering whether the sample induced a change of the original prediction within the NLI model. We further average the probability distribution of labels obtained from the final softmax layer of the model for these two categories.
We measure the differences between the two distributions with two statistical tests. To evaluate the relative entropy between them, we use Jensen-Shanon Divergence \citep{fuglede2004jensen}, a symmetric, non-negative, and bounded metric for assessing the similarity between two distributions, $\operatorname{JSD}(P \| Q)=\frac{1}{2} D(P \| M)+\frac{1}{2} D(Q \| M)$, where $D$ is the Kullback–Leibler divergence \citep{joyce2011kullback}. We verify the statistical significance of our findings with the Kolmogorov–Smirnov test \citep{berger2014kolmogorov}, which shows if the two sets of samples are likely to come from the same distribution.

Our results in \autoref{fig:div} show a significant distribution shift when assessing semantics-preserving surface-form variations. The cosine distance in the sentence embedding space between the generated and original samples is negligible at $0.04$. As the absolute cosine similarity values possess limited interpretable meaning, we further explore the distributions of cosine distances towards original samples for the examples that do and do not induce label changes. We measure the Jansen-Shannon divergence of these two distributions at $0.001$, implying they are strongly similar. This reinforces the hypothesis that surface-form variations produce logically equivalent samples with minor distance in the embedding space regardless of the induced label changes. 
However, despite minor changes in the semantic composition, we see a sizable change in the final predictive distribution of the NLI models. We see a significant rise both in Jensen-Shannon divergence and Kalmogorov-Smirnov metric, $\triangle \text{JSD} = 0.51$ and $\triangle \text{K-S} = 0.54$, when comparing the examples where the model prediction has changed compared to the original. This indicates that the generated variations do not cause negligible change within model prediction, but rather can be considered adversarial for the model. It shows that the limited capabilities to utilise syntactic information cause the model to significantly change the final prediction given minuscule variations, which is an to inconsistent predictive behaviour. Given that we initially sampled examples that the models answered correctly, these results assert our belief that the models do not display consistent predictive behaviour despite having equivalent inputs. This shows that albeit the strong model performance presented in \autoref{tab:nli_eval}, there is masked degeneration and discrepancies within the NLI models stemming from semantic sensitivity. Our method allows for explicitly quantifying the degree of semantic sensitivity within PLMs and allows to measure the impact of that sensitivity on the decision-making process of the model.

\paragraph{Semantic-Sensitivity and decision variations}

We 
lastly analyse the standard deviation within the predicted label distribution produced from the softmax of the model. We compute the standard deviation for the distribution of original premise hypothesis predictions and compare it with a replacement that does not and does cause label changes in PLM classification, see \autoref{fig:var}.
For reference, the upper bound for standard deviation in this 3 class setting happens when the model is greatly confident in one of the classes, i.e. $\text{softamx} = [1, 0, 0] \rightarrow \sigma_\text{max} = 0.471$. Bigger $\sigma$ on average implies more confident answers by the PLM. It can be observed that the average predictions with the original samples have a great degree of confidence. We see an interesting phenomenon where the predictive confidence slightly rises across most of the datasets for the cases where the model is able to recover the inherent textual relations. However, when faced with examples that cause label changes, there is a significant drop of $\triangle \sigma = 0.1$ in the standard deviation averaged across the datasets. This signifies that predictive confidence sizably degrades when the model struggles to recover the existent relations because of slight semantics-preserving variations. That further indicates that NLI models are susceptible to semantic sensitivity and have limited knowledge of compositional semantics, which can lead to the degradation of predictive confidence and incidentally inconsistent predictions.

\section{Conclusion}

We present a novel framework for assessing semantic sensitivity in NLI models through generating semantics-preserving variations. Our systematic study of the phenomenon across various datasets and transformer-based PLMs shows that the models consistently struggle with variations requiring knowledge of compositional semantics. This performance deterioration happens across the whole label space, almost regardless of model size. We measure the impact of semantic-sensitivity and show that it diminishes models' predictive confidence and can lead to predictive inconsistency.

\section*{Limitations}

In our work, we cover the semantic-sensitivity that can be found within NLI models. However, the framework can be applied to a wider range of classification tasks. The benchmark can be extended with more datasets and further enhanced with larger human evaluation. Also, we covered PLMs specifically trained for NLI; however, it would be great to cover bigger LLMs, in particular w.r.t. their emergent zero-shot capabilities. Another limitation is that we only cover English-based language models and do not test in multi-lingual or cross-lingual settings.

\section*{Ethics Statement}

Our work completes an analysis of numerous models w.r.t. their decision inconsistency induced by semantic surface form variations. We show that models are somewhat unable to handle logically and semantically equivalent sentences, which would lead to an inconsistent use across various domains and applications. Our generation method does not induce any further exploitation threat and can only be used for measuring the above-mentioned inconsistencies. We exclusively use open source publicly accessible data and models within our experimentations.

\section*{Acknowledgements}
Erik is partially funded by a DFF Sapere Aude research leader grant under grant agreement No 0171-00034B, as well as by a NEC PhD fellowship.
This work is further supported by the Pioneer Centre for AI, DNRF grant number P1.
%

\bibliography{custom}
\bibliographystyle{acl_natbib}

\clearpage
\appendix
\section{Appendix}
\label{sec:appendix}

\begin{table}[h]
\begin{tabular}{@{}lrrrr@{}}
\toprule
Dataset & \multicolumn{1}{l}{Fuzzy token match \%} & \multicolumn{1}{l}{average length $h$} & \multicolumn{1}{l}{average length $h\prime$} & \multicolumn{1}{l}{average token overlap} \\ \midrule
mnli & 84.83 & 14.31 & 14.14 & 13.25 \\
snli & 81.55 & 10.81 & 11.21 & 10.38 \\
anli\_r1 & 87.59 & 17.3 & 17.02 & 13.73 \\
anli\_r2 & 86.49 & 15.99 & 15.84 & 12.8 \\
anli\_r3 & 85.17 & 14.32 & 14.29 & 11.27 \\ \bottomrule
\end{tabular}
\caption{}
\label{tab:token_anal}
\end{table}

\paragraph{Evaluation under Label change}

\begin{figure}[h]
\includegraphics[width=\columnwidth]{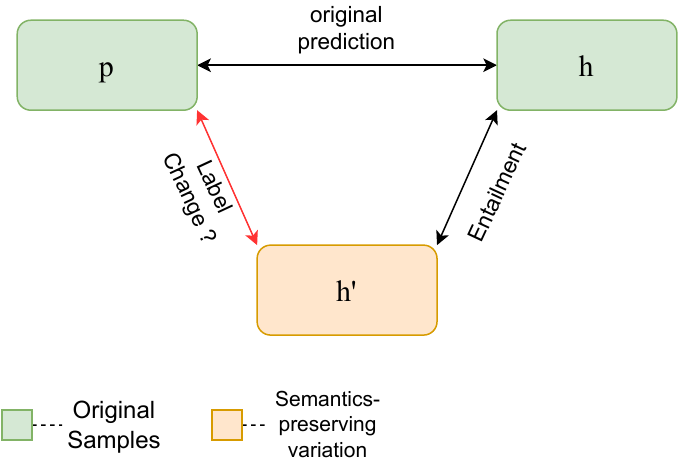}
    \caption{A diagram for assessing semantic similarity. Given the generated semantics-preserving surface-form variation $h^\prime$, we evaluate if a label change occurs when replacing the hypothesis in accordance with \autoref{eq:label_change}}.
    \label{fig:semantic_triangle}
\end{figure}

To assess the extent of the impact of semantic sensitivity, we employ an evaluation under label change. This means we consider the examples that changed the original prediction of the model after a surface-form variation replaced the original hypothesis. A graphical representation of this can be seen in \autoref{fig:semantic_triangle}. It must be noted that we use only the samples that the model originally predicted correctly to avoid incorrect assessment regarding the reasoning behind the false predictions. Our primary aim is to measure the semantic sensitivity within the model predictions and the extent of inconsistency it causes.

\paragraph{Token Level-Differences of the generated variations}

We further explore the difference between surface-form variations and original examples by conducting a token-level analysis for each pair $(h, h\prime)$. We compute the average amount of tokens present for the original and generated hypothesis and use fuzzy and exact matching to assess the overlap of tokens on average for each dataset. The results can be seen in \autoref{tab:token_anal}. The results show that the generated and original examples have a high token level overlap which further reinforces the idea that surface form variations are close both syntactically, in the embedding space and logically.

\end{document}

%% file: tables/nli_eval.tex
\begin{table*}[t]
\centering
\begin{tabular}{@{}c|cccccc@{}}
\toprule
         & \textbf{bart-l} & \textbf{roberta-l} & \textbf{distilbart} & \textbf{deberta-b} & \textbf{deberta-l} & \textbf{deberta-xl} \\ \midrule
MNLI$_{(n=10000)}$    & 90.10\%         & 90.56\%            & 87.17\%             & 88.77\%            & 91.32\%            & 91.44\%             \\ \midrule
SNLI$_{(n=10000)}$    & 87.55\%         & 86.44\%            & 84.37\%             & 84.39\%            & 88.87\%            & 88.54\%             \\
ANLI\_r1$_{(n=1000)}$& 46.20\%         & 46.40\%            & 41.40\%             & 35.10\%            & 49.70\%            & 53.00\%             \\
ANLI\_r2$_{(n=1000)}$& 31.60\%         & 27.00\%            & 32.80\%             & 29.80\%            & 32.70\%            & 35.40\%             \\
ANLI\_r3$_{(n=1200)}$& 33.08\%         & 26.75\%            & 32.75\%             & 30.50\%            & 35.92\%            & 38.75\%             \\ \bottomrule
\end{tabular}
\caption{The original accuracy on testing/dev sets for various transformers (b-base, l-large, xl-extra large) on \emph{in-domain} MNLI experiments and zero-shot transfers to \emph{out-of-domain} SNLI and ANLI. The number near the dataset name designates the exact amount of original samples in the testing set. }
\label{tab:nli_eval}
\end{table*}

%% file: tables/fooling_rate.tex

\begin{table*}[t]
\centering
\begin{adjustbox}{max width=\textwidth}
\begin{tabular}{@{}c|cccccc@{}}
\toprule
$r_s/r_r$   & \textbf{bart-large} & \textbf{roberta-large} & \textbf{distilbart} & \textbf{deberta-base} & \textbf{deberta-large} & \textbf{deberta-xlarge} \\ \midrule
MNLI      & 6.64\%/12.35\%                            & 5.71\%/11.56\%                               & 9.20\%/ \bf{16.80}\%                                  & 6.66\%/13.81\%                               & 5.38\%/11.54\%                               & 5.89\%/11.49\%                                \\ \midrule
SNLI      & 10.11\%/15.52\%                           & 8.38\%/14.98\%                               & 15.67\%/\bf{23.68}\%                               & 9.96\%/17.01\%                               & 7.83\%/13.39\%                               & 9.50\%/14.69\%                                 \\
ANLI\_r1  & 31.51\%/42.89\%                           & 28.45\%/35.01\%                              & 31.48\%/\bf{52.30}\%                                & 40.0\%/48.99\%                               & 25.66\%/37.88\%                              & 22.71\%/30.73\%                               \\
ANLI\_r2  & 34.39\%/51.91\%                           & 24.62\%/42.80\%                               & 36.09\%/\bf{57.49}\%                               & 34.92\%/48.47\%                              & 28.44\%/44.04\%                              & 29.46\%/46.46\%                               \\
ANLI\_r3  & 29.11\%/51.39\%                           & 21.88\%/45.00\%                               & 29.26\%/52.42\%                               & 33.88\%/\bf{53.17}\%                              & 24.88\%/44.65\%                              & 23.23\%/42.37\%                               \\ \bottomrule
\end{tabular}
\end{adjustbox}
\caption{The strict and relaxed fooling rates of different transformer models across \emph{in-domain} (MNLI) and \emph{out-of-domain} (SNLI, ANLI) evaluations. On average more than half of the labels change towards their logically contrasting counterpart.}
\label{tab:foolrate1}
\end{table*}

%% file: tables/fooling_rate_class.tex

\begin{table*}[t]
\centering
\begin{adjustbox}{max width=\textwidth}
\begin{tabular}{@{}c|cccc@{}}
\toprule
\textbf{} & $r_s/r_r\left(y=E\right)$ & $r_s/r_r\left(y=N\right)$ & $r_s/r_r\left(y=C\right)$ & $r_s/r_r$       \\ \midrule
MNLI                       & 2.78\%/13.41\%                     & 14.33\%/14.33\%                    & 3.69\%/11.17\%                     & 6.58\%/12.92\%     \\ \midrule
SNLI                       & 9.54\%/18.73\%                     & 19.42\%/19.42\%                    & 2.92\%/11.82\%                     & 10.24\%/16.54\%    \\
ANLI\_r1                   & 21.64\%/41.97\%                    & 38.62\%/38.62\%                    & 29.17\%/44.57\%                    & 29.97\%/41.30\%     \\
ANLI\_r2                   & 20.84\%/46.28\%                    & 49.41\%/49.41\%                    & 21.89\%/50.80\%                     & 31.32\%/48.53\%    \\
ANLI\_r3                   & 11.65\%/52.00\%                     & 47.18\%/47.18\%                    & 16.42\%/46.50\%                     & 27.04\%/48.17\%    \\ \bottomrule
\end{tabular}
\end{adjustbox}
\caption{Fooling rate averaged over all models. $r_s$ represents the strict fooling rate, in which case the predicted label of the evaluation pair is opposite to the original label $y$. $r_r$ measures the proportion of label change. $y\in\{E,N,C\}$ group the $(p,h)$ pairs by their semantic relation, representing entailment, neutrality, and contradiction, respectively.}
\label{tab:foolrate_class}
\end{table*}